\documentclass[sn-mathphys,Numbered]{sn-jnl}

\usepackage{graphicx}%
\usepackage{multirow}%
\usepackage{amsmath,amssymb,amsfonts}%
\usepackage{amsthm}%
\usepackage{mathrsfs}%
\usepackage[title]{appendix}%
\usepackage{xcolor}%
\usepackage{textcomp}%
\usepackage{manyfoot}%
\usepackage{booktabs}%
\usepackage{algorithm}%
\usepackage{algorithmicx}%
\usepackage{algpseudocode}%
\usepackage{listings}%

\usepackage[T1]{fontenc}

\begin{document}

\title[Article Title]{Runner re-identification from single-view running video in the open-world setting}

\author*[1]{\fnm{Tomohiro} \sur{Suzuki}}\email{suzuki.tomohiro@g.sp.m.is.nagoya-u.ac.jp}

\author[1]{\fnm{Kazushi} \sur{Tsutsui}}\email{tsutsui@g.sp.m.is.nagoya-u.ac.jp}

\author[1]{\fnm{Kazuya} \sur{Takeda}}\email{kazuya.takeda@nagoya-u.jp}

\author[1,2,3]{\fnm{Keisuke} \sur{Fujii}}\email{fujii@i.nagoya-u.ac.jp}

\affil*[1]{\orgdiv{Graduate School of Informatics}, \orgname{Nagoya University}, \orgaddress{\street{Chikusa-ku}, \city{Nagoya}, \state{Aichi}, \country{Japan}}}

\affil[2]{\orgdiv{RIKEN Center for Advanced Intelligence Project}, \orgname{1-5}, \orgaddress{\street{Yamadaoka}, \city{Suita}, \state{Osaka},  \country{Japan}}}

\affil[3]{\orgdiv{PRESTO}, \orgname{Japan Science and Technology Agency}, \orgaddress{\city{Kawaguchi}, \state{Saitama},\country{Japan}}}

\abstract{In many sports, player re-identification is crucial for automatic video processing and analysis.
However, most of the current studies on player re-identification in multi- or single-view sports videos focus on re-identification in the closed-world setting using labeled image dataset, and player re-identification in the open-world setting for automatic video analysis is not well developed.
In this paper, we propose a runner re-identification system that directly processes single-view video to address the open-world setting. In the open-world setting, we cannot use labeled dataset and have to process video directly.
The proposed system automatically processes raw video as input to identify runners, and it can identify runners even when they are framed out multiple times.
For the automatic processing, we first detect the runners in the video using the pre-trained YOLOv8 and the fine-tuned EfficientNet. We then track the runners using ByteTrack and detect their shoes with the fine-tuned YOLOv8. Finally, we extract the image features of the runners using an unsupervised method with the gated recurrent unit autoencoder and global and local features mixing. To improve the accuracy of runner re-identification, we use shoe images as local image features and dynamic features of running sequence images.
We evaluated the system on a running practice video dataset and showed that the proposed method identified runners with higher accuracy than some state-of-the-art models in unsupervised re-identification. We also showed that our proposed local image feature and running dynamic feature were effective for runner re-identification. Our runner re-identification system can be useful for the automatic analysis of running videos.
%
}


\maketitle

\section{Introduction}\label{Introduction}
In many sports, it is important to evaluate tactics and movements from videos, which allow us to collect and analyze useful information without interfering with the athlete's movements. Most of the previous studies in recent sports video processing have investigated methods for detecting and tracking players and balls \cite{Vandeghen_2022_CVPR_soccer_detect_player,Cioppa_2022_CVPR_soccer_tracking,scott2022soccerTrack,wang2022sportstrack}, and have mainly focused on ball games such as soccer, basketball, ice hockey, and rugby.
These approaches can be extended to detecting high-risk plays in rugby \cite{Nonaka_2022_CVPR_rugby} and offsides in soccer \cite{Uchida2021-bb-offside}, scoring in rhythmic gymnastics \cite{diaz2014automatic} and figure skating \cite{xu2019learning}, and detecting race-walk faults \cite{suzuki2022automatic,suzuki2023automatic}. 
For such applications for each athlete in videos, athlete detection, tracking, and re-identification are important topics for the automation of the processes.

\begin{figure}[ht]
    \centering
    \includegraphics[scale=0.5]{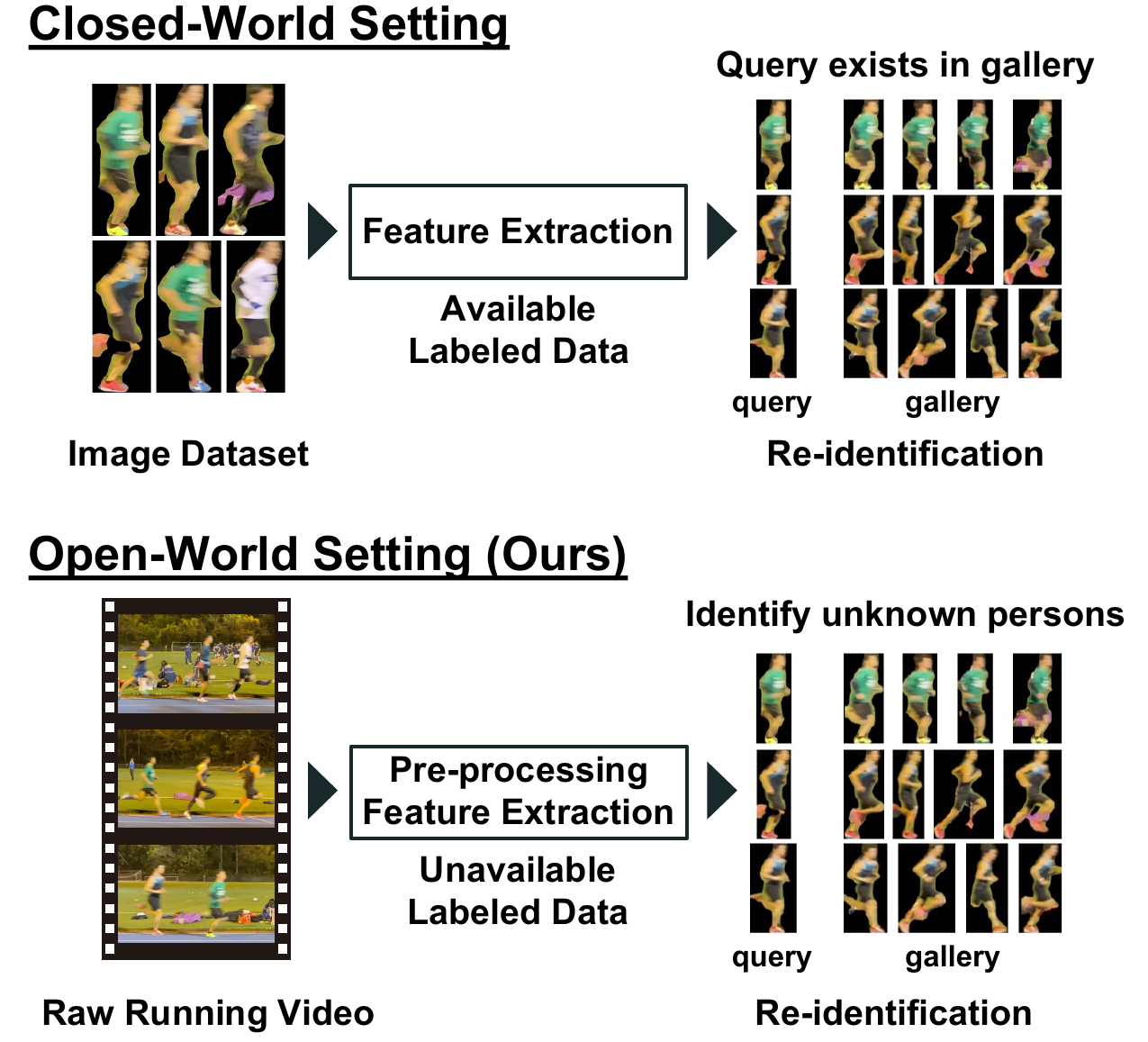}
    \caption{Overview of the two re-identification tasks.  Re-identification in the closed-world setting can use the labeled image dataset, but it is not directly applicable to real-world general sports video analysis. On the other hand, re-identification in the open-world setting, our proposed method, can directly process raw video and does not require labeled data for feature extraction.}
    \label{fig:overview}
\end{figure}

Player re-identification is the process of identifying each player in a video, which is crucial for automatic video processing in sports. Player re-identification uses the features of the player's image, such as jersey color and/or number \cite{Uchida2021-bb-offside,ivankovic2014automatic-basket-color,Napolean2019-oz-run-event}, or the feature vector obtained from the feature extractor \cite{Comandur2022-py,Koshkina2021-pn-hockey-classification,Vats2022-lq-hockey-tfidentification,Habel2022-ib-clip-basket,Napolean2019-oz-run-event}, to identify players in the multi- or single-video.
These studies are not directly applicable to general sports video processing because they are re-identification in the closed-world setting \cite{Ye2022-pd} (Figure \ref{fig:overview} Top), which focuses on improving the performance of the feature extractor using the prepared image dataset. In the closed-world setting, we can use the image dataset with sufficient labeled data to train the feature extractor. However, the cost of preparing the labeled image dataset from the video is enormous. In addition, real-world re-identification requires the identification of players who are not included in the dataset. Therefore, the closed-world setting is limited for player re-identification in general sports videos, and the open-world setting \cite{Ye2022-pd} (Figure \ref{fig:overview} Bottom) should be considered.
In the open-world setting, we need to directly process the raw image or video, including person image extraction and image feature extraction. It does not require manual data processing of the video. For these reasons, re-identification in the open-world setting is suitable for general sports video processing, especially daily practice videos.
For example, in daily practice, many non-professional athletes need to consider the cost of video measurement and processing with a fixed camera.
Our research motivation is to develop a runner re-identification system from single-view video in the open-world setting for daily practice use.

In track and field practice, especially in middle and long-distance running and race walking, runners make laps around the same place, such as a track, over and over again. In such situations, fixed-point video allows the runner to be captured many times. If only one runner appears in this video, for example, the runner's three-dimensional joint position information and motion dynamics can be analyzed \cite{Li2022-tw,uhlrich2022opencap}. However, these analysis techniques cannot distinguish between multiple runners in the video and analyze them simultaneously. To solve this problem, multi-object tracking can be used to identify runners as they pass through the capture area, but the tracking algorithm is unable to identify runners once they disappear from the screen. For this reason, runner re-identification in single-view videos captured at a fixed point plays an important role in the analysis of runner motion in practice videos.

\begin{figure}[h]
    \centering
    \includegraphics[width=\linewidth]{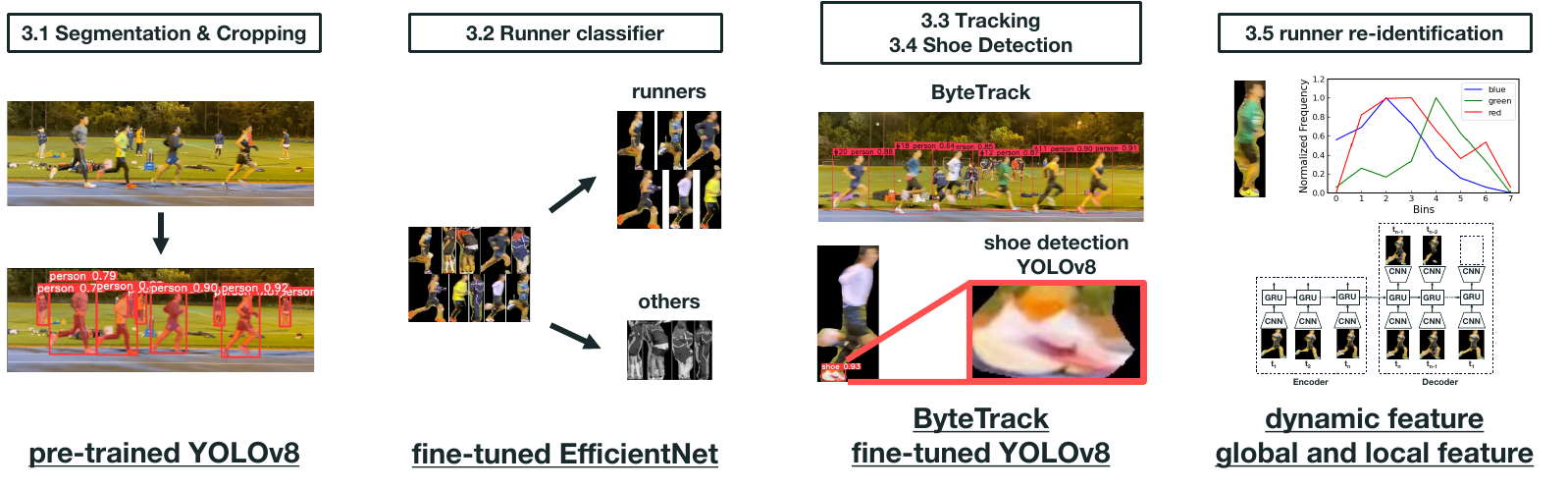}
    \vspace{-10pt}
    \caption{Our proposed runner re-identification system flow. In Section 3.1, we describe the segmentation and cropping model, and in Section 3.2, we describe the EfficientNet fine-tuned for runner classification. In Sections 3.3 and 3.4, we describe the tracking and the shoe detection separately. In Section 3.5, we describe the runner re-identification.}
    \label{fig:method}
    \vspace{-10pt}
\end{figure}

In this paper, we propose a runner re-identification system (see Figure \ref{fig:method}) that directly processes single-view video to address the open-world setting. In the system pipeline, we detect runners in the video using pre-trained YOLOv8 \cite{Jocher_YOLO_by_Ultralytics_2023} and fine-tuned EfficientNet \cite{tan2019efficientnet, tan2021efficientnetv2}. We then track them using Bytetrack \cite{zhang2022bytetrack} and detect their shoes using fine-tuned YOLOv8. Finally, we extract the runner's image features using the Gated Recurrent Unit AutoEncoder (GRU AE) and global and local mixed features extracted by Hard-sample Guided Hybrid Contrast Learning for Unsupervised Person Re-identification (HHCL) \cite{hu2021hard}. Our research aims to realize runner re-identification in the open-world setting using single-view running practice video as input and evaluate the results.
Since our method uses an unsupervised feature extraction method, it does not require the preparation of labeled data for feature extraction. In addition, our system consistently and automatically processes runner detection, tracking, and re-identification in the video. These are great advantages for the open-world setting.
We use a small amount of labeled data to fine-tune high-performance pre-processing models (runner classifier and shoe detector) required for the proposed system, such as runner detection and shoe detection. However, the labeled data is only needed for the initial fine-tuning of the pre-processing models, and once the pre-processing model is fine-tuned, the labeled data is no longer needed.

Our main contributions are:
\begin{enumerate}
   \item We propose a runner re-identification system from single-view video in the open-world setting, which supports automated video processing and analysis of daily practice. We evaluate the performance of the re-identification pipeline using raw video as input.
   \item We propose dynamic features and global and local mixing features obtained by unsupervised method such as GRU AE. Since the training of feature extraction models does not require labeled data, we can reduce the cost of automatic video processing.
   \item We show that our proposed method can identify runners in the practice videos with higher accuracy than some SOTA unsupervised re-identification methods. We also show that our proposed global and local feature mixing and dynamic features are effective for runner re-identification.
\end{enumerate}

This paper is organized as follows. First, in Section \ref{Related work}, we provide an overview of the related work. Next, we describe the pipeline of our proposed system in Section \ref{Methods}. Then, we explain our dataset and implementation details in Section \ref{Experiments}. Finally, we present the experimental results in Section \ref{Results} and conclude this paper in Section \ref{Conclusion}.

\section{Related work}\label{Related work}

To achieve runner re-identification from video in the open-world setting, we first detect runner passing scenes in the practice video, followed by runner re-identification between each detected scene. In other words, our research relates to the topics of sports video segmentation and player re-identification.

Segmentation of sports videos includes detecting specific actions (e.g., scoring and penalties) in the video \cite{Giancola2018-ih-soccernet,Cioppa2022-dv,Zhu2022-mu-tfaction-spotting,Uchida2021-bb-offside,Vats2020-wg-hockey-event}, detecting a series of actions \cite{McNally2019-ev-golfDB}, and extracting the active time in the game \cite{Pidaparthy2021-cu-hockey-seg,Napolean2019-oz-run-event}. In these approaches, the game situation is determined from the visual and optional auditory cues, or specific actions are detected by the movement of a single player. In contrast, detecting a runner's passing scene would be easier than in the above studies because runners in videos do not make complex movements as in other sports. For example, there is an example of detecting the passage of a runner by using text spotting to recognize the numbers in the image and matching them with the runner's ``bib number'' registered on the race event website \cite{Napolean2019-oz-run-event}. However, our task is challenging in terms of re-identification from raw videos in the open-world setting without bib numbers.

In player re-identification, many studies have used jersey color and number as features for re-identification \cite{Uchida2021-bb-offside,ivankovic2014automatic-basket-color,Napolean2019-oz-run-event}. For example, re-identification in running events \cite{Napolean2019-oz-run-event} uses the bib number of the runner in the race to identify each runner. Although the bib numbers are used during the race, they are not available to identify runners in the daily practice videos. In general, the colors of an athlete's clothing and shoes can be more common features than the numbers to characterize them.
However, since it is difficult to achieve high-performance re-identification with only these features, recent studies have used feature extractors based on metric learning or contrastive learning \cite{Comandur2022-py,Koshkina2021-pn-hockey-classification,Vats2022-lq-hockey-tfidentification,Habel2022-ib-clip-basket,Napolean2019-oz-run-event}. Furthermore, gait recognition \cite{Wan2018-ga,Chao2022-mq} is also effective for person re-identification, which captures unique features of gait dynamics. Such re-identification methods have contributed to the realization of high-performance player re-identification, but they have the problem of requiring large amounts of data and designing loss functions according to the data to be identified. In addition, many re-identification studies are the closed-world setting \cite{Ye2022-pd} that focuses on how to train effective feature extractors using player image datasets and does not consider the open-world setting \cite{Ye2022-pd} that identifies players directly from the video, which is important for automatic sports video analysis. Moreover, while some studies of re-identification for open-world settings have addressed unsupervised feature extraction methods \cite{ge2020selfpaced, Chen_2021_ICCV, hu2021hard, Cho2022-nq, Li2020-dh}, training of feature extractors with the small number of images that can realistically be acquired \cite{Zheng2016-hj}, and retrieval methods that can handle large amounts of data \cite{Zhu2018-bl}, no studies have evaluated pipeline systems for re-identification that process video directly.
To solve the above problems, we implement and evaluate a runner re-identification system that identifies runners directly from video, including pre-processing such as person detection. The feature extractor in our system does not require labeled data. Therefore, our system can directly identify unknown runners in the video.

\section{Methods}\label{Methods}

Our goal is to identify runners in a usual running practice video, even if the runners are framed out multiple times. To identify framed-out runners, we detect the running scenes for each runner in the video. Here we consider a ``running scene'' as a sequence of frames during which each runner enters the screen from the left and exits from the right. Videos of runners captured from the side are useful for analyzing the runner's form while running, including the trajectory of the runner's legs. Figure \ref{fig:method} shows the flow of our system in the following five steps.
First, persons in each frame of the videos are segmented. We employ the YOLOv8 \cite{Jocher_YOLO_by_Ultralytics_2023} segmentation model to segment persons. After the segmentation, each person's image is cropped for the next step.
Second, cropped images are classified as runners or not using fine-tuned EfficientNet \cite{tan2019efficientnet, tan2021efficientnetv2}.
Third, we track each runner using ByteTrack \cite{zhang2022bytetrack}.
Moreover, each runner's shoes are detected using the fine-tuned YOLOv8 shoe detector.
Finally, we identify the same runners from each tracked running scene. For the feature extractor, we use a Gated Recurrent Unit AutoEncoder (GRU AE) and global and local feature mixing by HHCL \cite{hu2021hard} as deep learning-based image features. In addition, we use simple color features, the RGB color histograms of the runner images. For comparison, we use four state-of-the-art (SOTA) unsupervised re-identification methods \cite{ge2020selfpaced, Chen_2021_ICCV, hu2021hard, Cho2022-nq} as baseline models. Since we only use labeled data for fine-tuning the runner classifier and shoe detector, our feature extractor is an unsupervised model.

\subsection{Person Segmentation and Cropping}
To process each person in each frame image individually, we segment and crop person images in each frame. We employ the segmentation model of YOLOv8 \cite{Jocher_YOLO_by_Ultralytics_2023} trained on the MS COCO dataset \cite{cocodataset} to segment persons. The segmented each person's image is cropped using bounding boxes and segmentation masks. The background color of the cropped image is undesired to identify runners. Therefore, we use instance segmentation instead of object detection. Note that, in the running video, the general background removal algorithm did not work well because the shapes of the runners were unclear due to motion blur.

\subsection{Runner Classifier}
The cropped images obtained from the above procedure include both runners and non-runners. Non-runner images should be removed because they would affect runner re-identification performance. Since the characteristics of runner images captured from the side are the same regardless of the video, a supervised approach does not compromise versatility. We train a runner classifier using labeled cropped person images. For the runner classification, we fine-tuned EfficientNet \cite{tan2019efficientnet, tan2021efficientnetv2} pre-trained on the ImageNet dataset \cite{deng2009imagenet}.

\subsection{Runner Tracking}
\begin{figure}[h]
    \centering
    \includegraphics[scale=0.45]{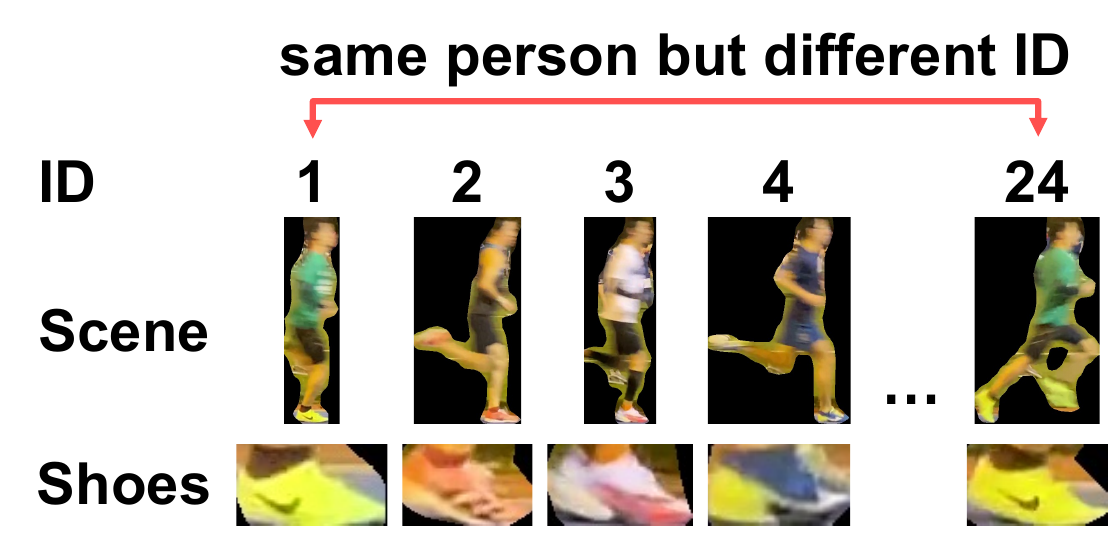}
    \caption{Example of tracking results. Representative images and a shoe image are obtained for each runner. Note that different IDs are assigned to the same person who reappears on the screen.}
    \label{fig:tracking}
    \vspace{-5pt}
\end{figure}

In the tracking step, a unique ID is assigned to every runner in each running scene.
We employ ByteTrack \cite{zhang2022bytetrack} for runner tracking.
The coordinates of the runner's bounding box obtained by the runner classifier are input to ByteTrack. Then, the unique ID of each runner in each running scene is obtained. Note that in this step, the same runner has a different ID for each running scene, as shown in Figure \ref{fig:tracking}, because the tracking algorithm considers the same person reappearing on the screen as a different person.

During tracking, continuous images of each runner's motion for two steps are stored with an ID.
We used the correlation between the running motion periods and the width of the runner images to unify the running motion periods of the continuous images. The GRU AE is expected to extract the unique dynamic features of each runner's movements by unifying the period of the running movements.

\subsection{Shoe Detection}
In general, it is difficult to identify different runners wearing similar color clothes using only whole-body representative images.
To address this issue, we detect the shoes of runners, one of the characteristic equipment for them, as a local image feature for runner re-identification.

In order to detect shoes, we employ a fine-tuned YOLOv8 object detection model. We used manually annotated images of runners whose shoes were clearly recognized for training. Figure \ref{fig:shoes_detect} shows examples of shoes that are "clear" and "unclear".
When multiple shoes are detected, the one with the highest confidence score is stored with each runner ID for the following runner re-identification.

\subsection{Runner Re-Identification}
In runner re-identification, we identify runners based on the similarity of various image features. We use the deep learning-based image feature vector and the color features of the image as image features.

\begin{figure}[h]
    \centering
    \includegraphics[scale=0.4]{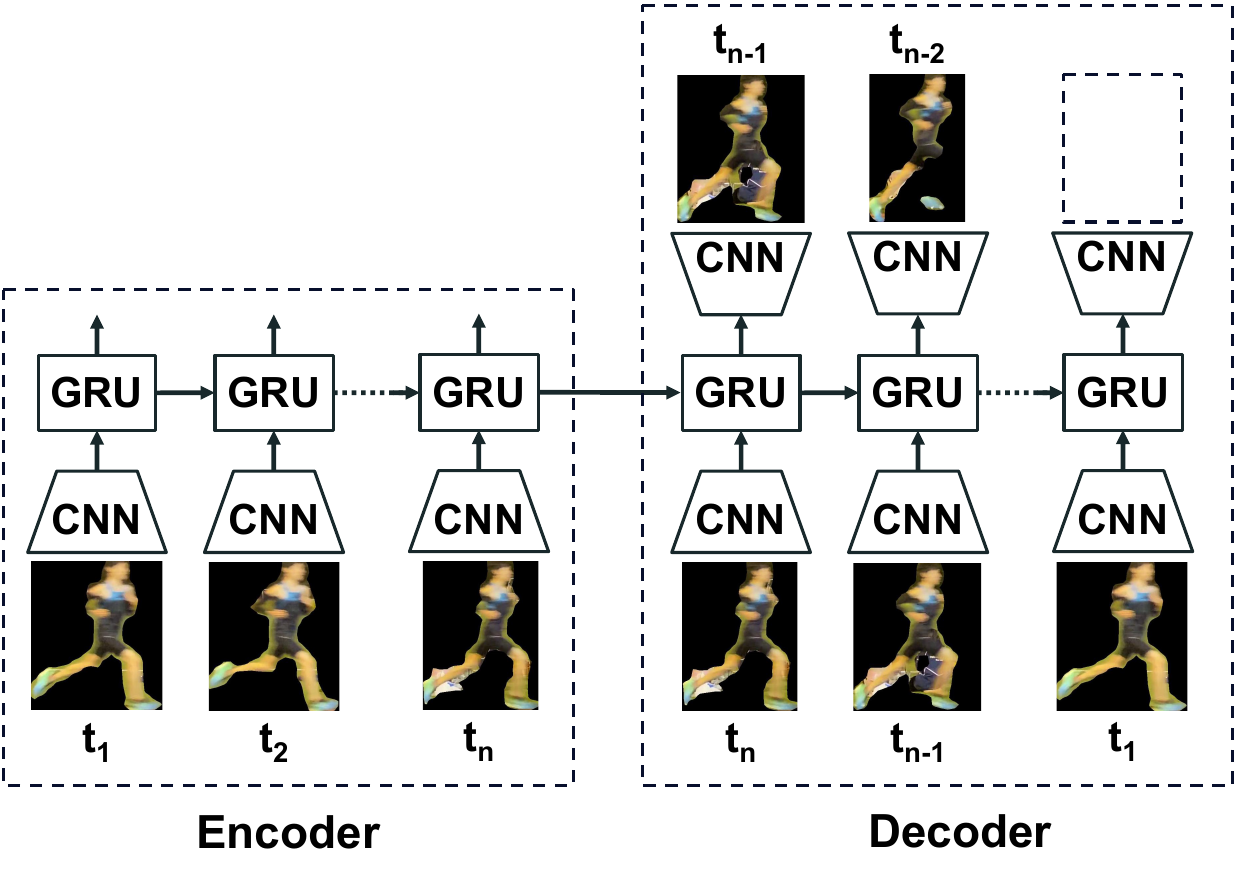}
    \caption{Model structure of the GRU AE. The encoder receives sequential images of running motion and outputs a 128-dimensional latent variable. The decoder receives latent variable and inverse-ordered sequential images and outputs the reconstructed image one frame before each input image.}
    \label{fig:gruae}
    \vspace{-5pt}
\end{figure}

To obtain a deep learning-based image feature vector, we propose a GRU AE that aims to acquire dynamic features from sequential images of running motion. Figure \ref{fig:gruae} shows the model structure of the GRU AE. We refer to the model structure of the sequence to sequence autoencoder \cite{Srivastava2015-ru}. The encoder of the model transforms the sequence images of running motion into vectors using CNN, which are then fed into the GRU in time series order. The last hidden layer of the GRU is then output to the decoder as a latent variable. The decoder receives the latent variables and the time series images in reverse order and reconstructs the image one frame before the input image by deconvolution of the output of the GRU. In the re-identification step, the encoder is used as a feature extractor, and latent variables are used as feature vectors of the image.

In addition to the GRU AE, we propose a combination of runner and shoe features using HHCL \cite{hu2021hard}. HHCL \cite{hu2021hard} is a high-performance unsupervised re-identification method that efficiently learns hard samples (i.e., images that are difficult to identify). The normal HHCL is trained unsupervised using person images. However, in this study, we also train the model using shoe images with the HHCL method. Although this study uses shoes as a local feature of the runner's image, the idea of using local features of the image could be extended to the other re-identification study. We use features from both models trained on runner images and models trained on shoe images for re-identification.

For the color features, the RGB color histograms obtained from the runner images and shoe images are used to identify the runner. The color histograms are obtained by dividing 8 bins for each channel. Note that we ignore the black pixel (background color) when calculating the histograms.
The histograms of each channel were flattened to obtain 24 dimensional ($=3 $ channels $\times 8$ bins) color features. For the runner images, since the lower body color features are similar for all runners, only the upper body color features are used.

After feature extraction, the similarity between each feature is calculated. We use it to evaluate the performance of re-identification. The similarity between the deep learning-based feature vectors is defined as the cosine similarity. The similarity between color features is defined as the correlation of histograms. There are two types of color similarity: the similarity of the histogram of the upper body image and the similarity of the histogram of the shoe image. When combining the above two color similarities, $90\%$ of the color similarity of the upper body image and $10\%$ of the color similarity of the shoe image are summed. Furthermore, the combined similarity of GRU AE features and color features is defined as follows:
\begin{equation}
\label{eq:sim}
  sim_{comb} = \lambda sim_{g} + (1 - \lambda) sim_{c},
\end{equation}
where $l$, $c$ denote GRU AE and color, $sim$ denotes similarity, and $\lambda$ denotes the hyper-parameter. 

In the similarity calculation, we introduced lap time filtering as domain knowledge from running training. Once the runners pass in front of the camera, they do not reappear for a while. Therefore, the similarity with the features extracted from running scenes that satisfy the conditions of the following equation is calculated as zero.
\begin{equation}
\label{eq:lap}
|S_{c_i} - S_q| > th,
\end{equation}
where $c_i$ denotes one of the running scenes $i$, $q$ denotes the query, $S$ denotes the start frame of the running scene, and $th$ is the threshold value of the lap time. Since it depends on situations or prior knowledge, here we set $th$ to 3,600 frames ($60$ s).

\section{Experiments}\label{Experiments}

\subsection{Dataset}
To build our dataset for the experiments, we captured long-distance running practice videos on the track. The videos were captured at 4K and 60 fps, using an iPhone 11 Pro set up to capture the runner from the side. In each lap, the runners passed through a section of approximately 10 meters, which is the camera's capturing range. Since we were capturing regular practice videos, there were a lot of non-runners, such as other athletes and staff, in the videos.

We prepared three datasets for the experiments. The first dataset is for fine-tuning pre-processing models such as the runner classifier (EfficientNet) and shoe detector (YOLOv8) in the runner re-identification system. For this dataset, we used two videos of 10 minutes each, taken during daytime and nighttime. We cropped each person's image in the videos and annotated whether each person was a runner or not. We also annotated some of the runner images with bounding boxes of running shoes. As a result, we obtained 5,000 images of runners and 12,500 images of people who are not runners. In addition, 600 images of runners were annotated with shoe bounding boxes.

The second dataset is a running scene image dataset used to train GRU AE for unsupervised feature extraction. A running sequence is a series of images of each runner's running motion for two steps extracted from each running scene. We extracted running sequence images from a total of six videos, three each for daytime and nighttime, using the proposed system. As a result, we obtained 1,182 sets of running sequence images for daytime and 1,114 sets of running sequence images for nighttime. Some of these data were also used to train the baseline models \cite{ge2020selfpaced, Chen_2021_ICCV, hu2021hard, Cho2022-nq} (see Section \ref{ImpPPLR} for details). Since the GRU AE and baseline models can be trained using an unsupervised method, no annotation is required for this dataset.

The third dataset was used to evaluate the performance of the proposed runner re-identification system, and was generated from 20 minutes of video for each daytime and nighttime. We extracted the sequence images of each running scene from these videos using the proposed system. We then annotated a unique runner ID for each running scene. Table \ref{tab:dataset} shows the number of running scenes and unique runners in each video obtained by annotation.

\begin{table}[h]
  \caption{The number of running scenes and unique runners in each video.}
  \label{tab:dataset}
  \begin{tabular}{lcc}
    \toprule
    Video type & running scenes & unique runners\\
    \midrule
    Daytime & 331 & 30\\
    Nighttime & 330 & 26\\
  \bottomrule
\end{tabular}
\end{table}

Note that the annotation data was only used for fine-tuning the runner classifier and the shoe detector, and for evaluating the proposed system. We did not use the annotation data for training the re-identification model to address the open-world setting.
While runner image datasets from race events exist \cite{penate2020tgc20reid, Napolean2019-oz-run-event}, to the best of our knowledge, this is the only dataset for runner re-identification for daily practice. Our dataset does not contain features that can easily identify runners, such as bib numbers used in races. Therefore, runner re-identification in this paper would be more difficult than the race event dataset \cite{Napolean2019-oz-run-event}, even though it is in the closed-world setting (note that our task is re-identification in the open-world setting).

\subsection{Implementation and Evaluation}
We evaluated the performance of the runner classifier, the shoe detector, and the runner re-identification. Our system was implemented in Python 3 with Pytorch.
Our code, evaluation dataset, and demo video will be available at \url{https://github.com/SZucchini/runner-reid}.

For the fine-tuning of the runner classifier, 14,000 annotated images were used as training data and 3,500 images as test data. In addition, $20\%$ of the training data was used as validation data.
The number of runner and non-runner images in the training data was 4000 and 10000, while the number of runner and non-runner images in the test data was 1000 and 2500.
We trained the model with a binary cross entropy loss function, employing the Adam optimizer with a learning rate of $3 \times 10^{-5}$. Due to the bias in the number of data for runners and non-runners, the $\textrm{F1-score}$ was used to evaluate the classification performance. 

For the fine-tuning of the shoe detector, 500 of the images of the runners annotated with shoe bounding boxes were used for training and 100 for testing.
We fine-tuned two different pre-trained models of YOLOv8 (yolov8n and yolov8m) and chose the one that showed better detection performance on the test data (yolov8n). The training setting of the fine-tuning was the same as the pre-trained model. Detection performance was evaluated using mean average precision (mAP$_{50\textrm{-}95}$) on the test data.

For runner re-identification, we compare the performance of four high-performance unsupervised re-identification models \cite{ge2020selfpaced, Chen_2021_ICCV, hu2021hard, Cho2022-nq} (as baseline models) and five proposed feature extractors using mean average precision (mAP) and Cumulative Matching Characteristic (CMC) rank-n accuracy. These evaluation metrics are used in many re-identification studies \cite{Comandur2022-py,Habel2022-ib-clip-basket,Wieczorek2021OnTU, ge2020selfpaced, Chen_2021_ICCV, hu2021hard, Cho2022-nq,Ye2022-pd}. Since our dataset is smaller than the major re-identification dataset \cite{market, msmt, veri}, we use $n=1,5$ in the rank-n accuracy. The feature extractors used to compare performance are baseline models \cite{ge2020selfpaced, Chen_2021_ICCV, hu2021hard, Cho2022-nq}, color histograms of upper body images (Color hist. w/o shoes), color histograms of upper body images and shoe images (Color hist. w/ shoes), autoencoder with input sequence images of running (GRU AE w/o color hist.), a combination of autoencoder and color histogram (GRU AE w/ color hist.), and global and local fetures mixing by HHCL (HHCL w/ shoes). The details of each feature extractor are described below.

\subsubsection{Baseline models}\label{ImpPPLR}
We used some SOTA unsupervised re-identification models \cite{ge2020selfpaced, Chen_2021_ICCV, hu2021hard, Cho2022-nq}, which can extract effective image features with unsupervised learning. These model perform well on some re-identification datasets such as Market-1501 \cite{market}, MSMT17 \cite{msmt}, and VeRi-776 \cite{veri}. In the experiments, we used models trained on our running scene images dataset as a baseline. Since we could not use all images from the dataset due to GPU memory limitations, we used 12,059 runner images from the daytime dataset and 10,229 runner images from the nighttime dataset, respectively. The models were trained separately for each dataset. In the re-identification performance evaluation phase, running scene images from the evaluation dataset were input to the model, and the 2,048-dimensional feature vectors from each image were averaged for each sequence. The cosine similarity of each averaged vector was then calculated as the similarity between each running scene.

\subsubsection{Color histogram without shoes}
We used only the RGB color histograms of the runner's upper body images as the image feature vector. We resized all images to $(W, H) = (64, 64)$ and defined the upper half of the image as the upper body.

\subsubsection{Color histogram with shoes}
In addition to the upper body image histograms, the RGB color histograms of each runner's shoe images were used as feature vectors. Each shoe image was resized to $(W, H) = (200, 100)$. The similarity of the upper body vectors and the shoe vectors was calculated separately. We averaged the body color similarity and the shoe color similarity after calculation.

\subsubsection{GRU AE without color histogram}
We trained the GRU AE using the running scene images dataset. Since the runner in the video looks very different during daytime and nighttime, we trained different models separately to evaluate each video.
In both cases, we used $80\%$ of the image sets for training and the rest for validation. We trained the model with mean square error as the loss function, employing the Adam optimizer with a learning rate of $1 \times 10^{-3}$. The dimension of the latent variable was 128.

\subsubsection{GRU AE with color histogram}
We combined the similarity of the image feature vectors in Color w/ shoes with that in GRU AE according to Equation (\ref{eq:sim}). We used mAP as a comparison metric between different lambda. Then we chose a lambda that gives the highest mAP as $0.85$.

\subsubsection{HHCL w/ shoes}
We trained HHCL \cite{hu2021hard}, the SOTA unsupervised re-identification model, on images of runners and shoes, respectively. In other words, we obtained a model that identifies the runner and a model that identifies the shoe separately. The similarity calculated from the features of each model between the runner images and the shoe images was then weighted and averaged at a ratio of $3:1$, and this similarity was used for re-identification.

\section{Results}\label{Results}

In this section, we present the results of the performance of the runner classifier, the shoe detector, and the runner re-identification.

\begin{figure}[h]
    \centering
    \includegraphics[scale=0.35]{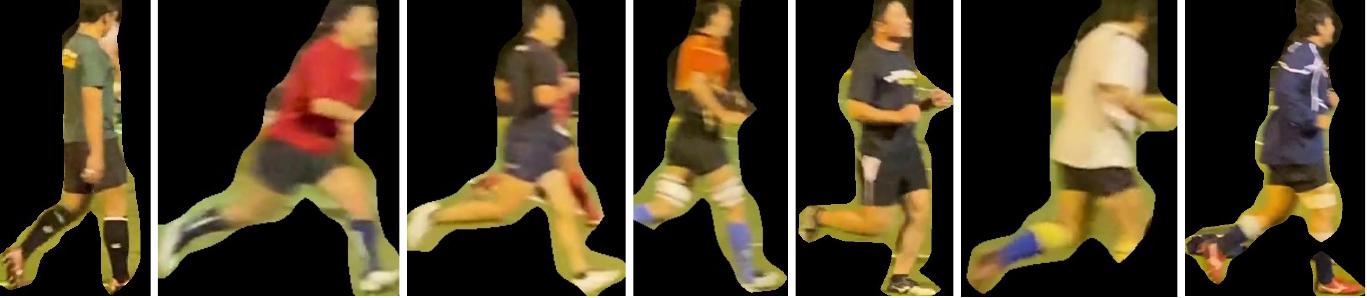}
    \caption{Examples of images misclassified as runners. Although we accomplished a $99.8\%$ \textrm{F1-score}, most of the misclassifications were images of players from other athletes running.}
    \label{fig:misclassification}
\end{figure}

For the runner classifier, the $\textrm{F1-score}$ was $99.8\%$. An example of images misclassified as runners is shown in Figure \ref{fig:misclassification}. Most of the misclassified images were of players running in other sports. Misclassification was addressed by filtering the results after runner tracking using frame length and image shape. Filtering resulted in the removal of approximately $98\%$ of non-runner images and incomplete runner images not suitable for re-identification.

\begin{figure}[h]
    \centering
    \includegraphics[scale=0.5]{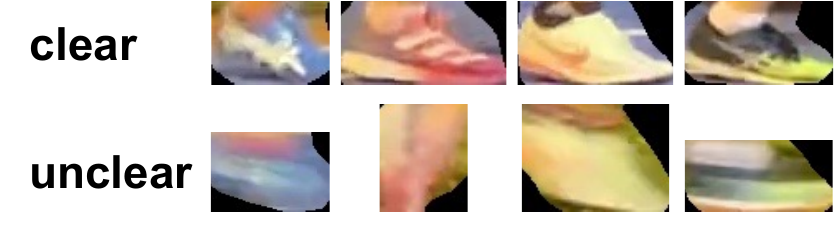}
    \caption{The shoe detection results. Although we accomplished a higher detection performance (upper), the model sometimes detected unclear shoes (lower).}
    \label{fig:shoes_detect}
\end{figure}

For the shoe detector performance, the mAP$_{50\textrm{-}95}$ was $86.2\%$. As an example of an error, the system detected unclear shoes which is difficult to use to compare the color features for re-identification, as shown in Figure \ref{fig:shoes_detect}. However, since the pre-trained model without fine-tuning could not detect ``shoe'', which are objects defined in the COCO dataset, fine-tuning was effective for shoe detection in our research.

For the runner re-identification, Table \ref{tab:mAP} shows the mAP and Rank-n accuracy for each model.
In the evaluation for daytime video, the proposed method ``GRU AE w/ hist.'' outperformed all the high performance re-identification methods in the previous studies except HHCL. For daytime videos, the images of runners obtained by segmentation are clearer and dynamic features are considered easier to extract. In contrast, in the evaluation for night videos, the performance of ``GRU AE w/o hist.'' is significantly lower, indicating that dynamic features are not well extracted.
Since the image quality of nighttime video is poor, motion blur appears in the images, and the segmentation accuracy of the runner image extraction is decreased. This makes it difficult for the proposed method to extract the color and dynamic features of the runner images, resulting in poor performance. For this reason, our proposed method may be unstable to changes in the video recording environment compared to previous studies.
Based on the above results, we conclude that our proposed dynamic features of running motion are effective in daytime conditions when the images are clear.

While ``GRU AE w/ hist.'' performed well for daytime videos, it could not outperform HHCL, a method that effectively uses hard samples for training. Since the runner dataset in our study has a lot of data from the same runner and many combinations of hard samples, we consider that HHCL was effective. Therefore, our proposed ``HHCL w/ shoes'', which learns HHCL on runner and shoe images and combines their features, also works well. ``HHCL w/ shoes'' outperformed all of the proposed and comparative methods, while also showing stable results for both daytime and nighttime videos. This method successfully utilizes the shoe image, a local feature in the image, to achieve more accurate re-identification. The combination of global and local image features is also considered to be effective for general re-identification.


\begin{table}[h]
  \caption{Comparison of the re-identification performance with feature extraction models.}
  \label{tab:mAP}
  \begin{tabular}{l|ccc|ccc}
    \toprule
    \multirow{2}{*}{Method} & \multicolumn{3}{c|}{Daytime} & \multicolumn{3}{c}{Nighttime} \\
    & mAP & Rank-1 & Rank-5 & mAP & Rank-1 & Rank-5\\
    \midrule
    SpCL \cite{ge2020selfpaced} & 0.874 & 0.954 & 0.985 & 0.884 & 0.971 & 0.974\\
    ICE \cite{Chen_2021_ICCV} & 0.869 & 0.960 & 0.978 & 0.863 & 0.965 & 0.971\\
    HHCL \cite{hu2021hard} & 0.940 & \textbf{0.985} & \textbf{0.988} & 0.911 & \textbf{0.974} & \textbf{0.981}\\
    PPLR \cite{Cho2022-nq} & 0.810 & 0.957 & \textbf{0.988} & 0.785 & 0.968 & 0.974\\
    Color hist. w/o shoes & 0.765 & 0.920 & 0.954 & 0.664 & 0.904 & 0.942\\
    Color hist. w/ shoes & 0.830 & 0.938 & 0.963 & 0.761 & 0.923 & 0.955\\
    GRU AE w/o color hist. & 0.852 & 0.957 & 0.981 & 0.636 & 0.875 & 0.958\\
    GRU AE w/ color hist. & 0.900 & 0.975 & \textbf{0.988} & 0.788 & 0.946 & 0.968\\
    HHCL w/ shoes & \textbf{0.944} & \textbf{0.985} & \textbf{0.988} & \textbf{0.915} & \textbf{0.974} & 0.978\\
  \bottomrule
\end{tabular}
\end{table}





\begin{figure}[t]
    \centering
    \includegraphics[width=0.85\linewidth]{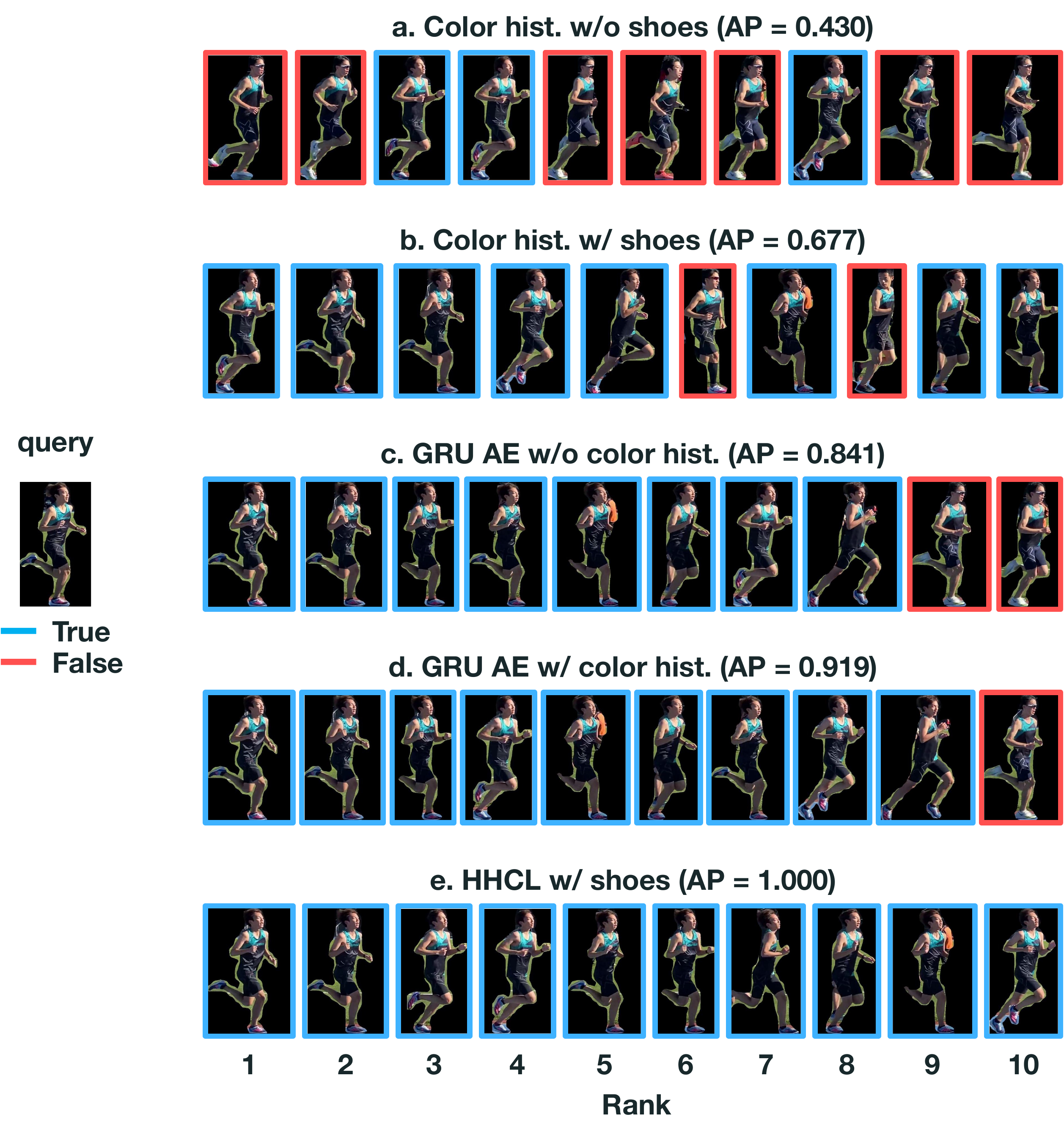}
    \caption{Re-identification result examples of our proposed methods. In each row, the similarity is higher for the left image. Runners identical to the query are marked with blue boxes and different runners are marked with red boxes. \textbf{(a)} ``Color hist. w/o shoes'' cannot identify runners wearing clothing similar to the query. \textbf{(b)} While an improvement over (a), ``Color hist. w/ shoes'' cannot identify runners wearing shoes similar to the query. \textbf{(c)} ``GRU AE w/o color hist.'' can identify runners with higher similarity than (b). \textbf{(d)} ``GRU AE w/ color hist.'' provides more accurate identification by combining dynamic and color features. \textbf{(e)} ``HHCL w/ shoes'' had all images of the same runner with the example query ranked higher in similarity than the images of the other runners.}
    \label{fig:comp}
\end{figure}

To evaluate the effectiveness of our proposed methods, we show a typical example of the re-identification results. Figure \ref{fig:comp} shows the top 10 similarity runner images for each of our proposed methods for an example query.
First, in Figure \ref{fig:comp}a, since only the upper body color feature is used for re-identification, 7 of the top 10 similarity runners are different runners from the query wearing the same clothing as the query.
In contrast, in Figure \ref{fig:comp}b, the combination of clothing and shoe color features works effectively, and 8 of the top 10 similarity runners are the same runner as the query. On the other hand, misidentified runners wearing shoes similar to the query are ranked in the top 10 in similarity, such as 6th and 8th similarity.
In Figure \ref{fig:comp}c, the top 8 runners are the same as the query. This result is more accurate than the results using color features (a and b), and the higher Average Precision (AP) shows that more accurate results are obtained for images outside the similarity range shown in the figure. This indicates that GRU AE's dynamic features other than appearance are more effective. 
Moreover, in Figure \ref{fig:comp}d, the combination of the GRU AE features and the color features results in the top 9 out of the top 10 similarities being the same runners as the query.
Finally, in Figure \ref{fig:comp}e, the combination of runner and shoe image features from each trained HHCL model (``HHCL w/ shoes'') results in all runners in the top 10 being correct. In addition, ``HHCL w/ shoes'' has all images of the same runner with the example query ranked higher in similarity than the images of the other runners ($AP=1.0$).

From the above results, the shoes, which are the unique equipment of runners, would be effective for re-identification. In addition, in the case of daytime video with clear images, the dynamic features extracted by GRU AE functioned as more effective features than the color features. While some runners wear similar clothes and shoes, the dynamic features of the runners (i.e., their running form) are unique to each runner and could be an important feature for runner re-identification.
We also found that HHCL is more effective for datasets with many combinations of hard samples.

\section{Conclusion}\label{Conclusion}
In this paper, we proposed and evaluated a runner re-identification system for a single-view video in the open-world setting. The evaluation results showed that our proposed method could identify runners in running practice videos with higher accuracy than some high performance unsupervised re-identification methods. We also showed that extracting dynamic features of running using the GRU AE and runner-specific local features of the image (e.g., shoes) were effective for runner re-identification in the single-view video. Our runner re-identification system can be useful for the automatic analysis of running videos.

However, we found that the proposed GRU AE was not robust against differences in brightness during video recording. In some cases, it was not possible to distinguish between runners with similar appearances, even when GRU AE dynamic features were used. For future work, a better pre-processing model, such as one that allows accurate runner segmentation exploiting novel segmentation methods \cite{Qin2023-yj, Liang2023-rs, yan2022solve}, could be useful to address this issue. Besides, features that are not affected by appearance, such as joint location information, can be effective. In addition, although the GRU AE in this study used a simple model architecture combining CNN and GRU, we consider that more accurate re-identification could be achieved by using more sophisticated representation learning methods such as space-time correspondance learning methods \cite{Qin2023-sn, Qin2023-zk}, Transformer-based autoencoder \cite{he2022masked}, or video autoencoder \cite{tong2022videomae} with high-memory GPUs.

\section*{Acknowledgments}
This work was financially supported by JSPS Grant Number 20H04075 and JST PRESTO Grant Number JPMJPR20CA.

\section*{Declarations}
\subsection*{Data availability statements}
The evaluation datasets generated during and/or analyzed during the current study will be available at \url{https://github.com/SZucchini/runner-reid}. However, data from participants who only agreed to participate in this study and did not consent to the release of their data may not be released.

\subsection*{Compliance with Ethical Standards}
The participants were fully informed about the study and their consent was obtained in advance. All the experimental procedures were performed after obtaining prior approval for ``Experiments on Human Subjects'' from the Graduate School of Informatics, Nagoya University.

\bibliography{sn-bibliography}

\end{document}